\documentclass[iicol,sn-mathphys,Numbered]{sn-jnl}% Math and Physical Sciences Reference Style
\usepackage{graphicx}%
\usepackage{multirow}%
\usepackage{amsmath,amssymb,amsfonts}%
\usepackage{amsthm}%
\usepackage{mathrsfs}%
\usepackage[title]{appendix}%
\usepackage{xcolor}%
\usepackage{textcomp}%
\usepackage{manyfoot}%
\usepackage{booktabs}%
\usepackage{algorithm}%
\usepackage{algorithmicx}%
\usepackage{algpseudocode}%
\usepackage{listings}%
\usepackage{tabularx}

\usepackage{adjustbox}
\usepackage{caption}
\theoremstyle{thmstyleone}%
\theoremstyle{thmstyletwo}%

\theoremstyle{thmstylethree}%

\begin{document}

\title[Article Title]{Representation Alignment Contrastive Regularization for Multi-Object Tracking}


\author[1,3]{\fnm{Zhonglin} \sur{Liu}}\email{zhonglinliu0@outlook.com}

\author*[1,3]{\fnm{Shujie} \sur{Chen}}\email{chenshujie@zjgsu.edu.cn}

\author[1,3]{\fnm{Jianfeng} \sur{Dong}}\email{dongjf24@gmail.com}

\author[1,3]{\fnm{Xun} \sur{Wang}}\email{wx@mail.zjgsu.edu.cn}

\author[2]{\fnm{Di} \sur{Zhou}}\email{zhoudi@uniview.com}

\affil*[1]{\orgdiv{College of Computer Science and Technology}, \orgname{Zhejiang Gongshang University}, \orgaddress{\city{Hangzhou}, \postcode{310018}, \country{China}}}

\affil[2]{\orgname{Zhejiang Uniview Technologies Co.,Ltd.}, \orgaddress{\city{Hangzhou}, \postcode{310051}, \country{China}}}

\affil[3]{\orgname{Zhejiang Key Laboratory of Big Data and Future E-Commerce Technology}, \orgaddress{\city{Hangzhou}, \postcode{310018}, \country{China}}}


\abstract{Achieving high-performance in multi-object tracking algorithms heavily relies on modeling spatio-temporal relationships during the data association stage. Mainstream approaches encompass rule-based and deep learning-based methods for spatio-temporal relationship modeling. While the former relies on physical motion laws, offering wider applicability but yielding suboptimal results for complex object movements, the latter, though achieving high-performance, lacks interpretability and involves complex module designs. This work aims to simplify deep learning-based spatio-temporal relationship models and introduce interpretability into features for data association. Specifically, a lightweight single-layer transformer encoder is utilized to model spatio-temporal relationships. To make features more interpretative, two contrastive regularization losses based on representation alignment are proposed, derived from spatio-temporal consistency rules. By applying weighted summation to affinity matrices, the aligned features can seamlessly integrate into the data association stage of the original tracking workflow. Experimental results showcase that our model enhances the majority of existing tracking networks' performance without excessive complexity, with minimal increase in training overhead and nearly negligible computational and storage costs. Our code is available at \href{https://github.com/liuzhonglincc/RATracker}{https://github.com/liuzhonglincc/RATracker}.}


\maketitle

\section{Introduction}\label{sec:introduction}
Multi-Object Tracking (MOT) has been a long-standing challenge in the field of computer vision \cite{wojke2017simple, feichtenhofer2017detect, bergmann2019tracking}. The main objective of MOT is to accurately determine the positions of various objects of interest within a video and to establish distinct trajectories for each of these objects. The potential applications of high-resolution MOT are widespread, encompassing areas such as autonomous driving \cite{fremont2020formal}, video analysis \cite{arnab2021vivit, liu2022video}, and scene comprehension \cite{sharir2021image}.

While many researchers \cite{wang2020towards, zhang2021fairmot, zhou2020tracking, peng2020chained} are increasingly inclined towards addressing the MOT problem by simultaneously tackling both object detection and tracking, the tracking-by-detection (TBD) approach remains a prominent paradigm in MOT due to its efficiency and cost-effectiveness \cite{tang2017multiple, xu2019spatial, zhang2022bytetrack}. In the tracking-by-detection approach, the MOT task is divided into two distinct tasks: object detection and association. The first task involves identifying and localizing target objects in each frame, while the second task revolves around solving the challenge of associating historical trajectories with presently detected objects.

\begin{figure}[tb]
    \centering
    \includegraphics[width=1\columnwidth]{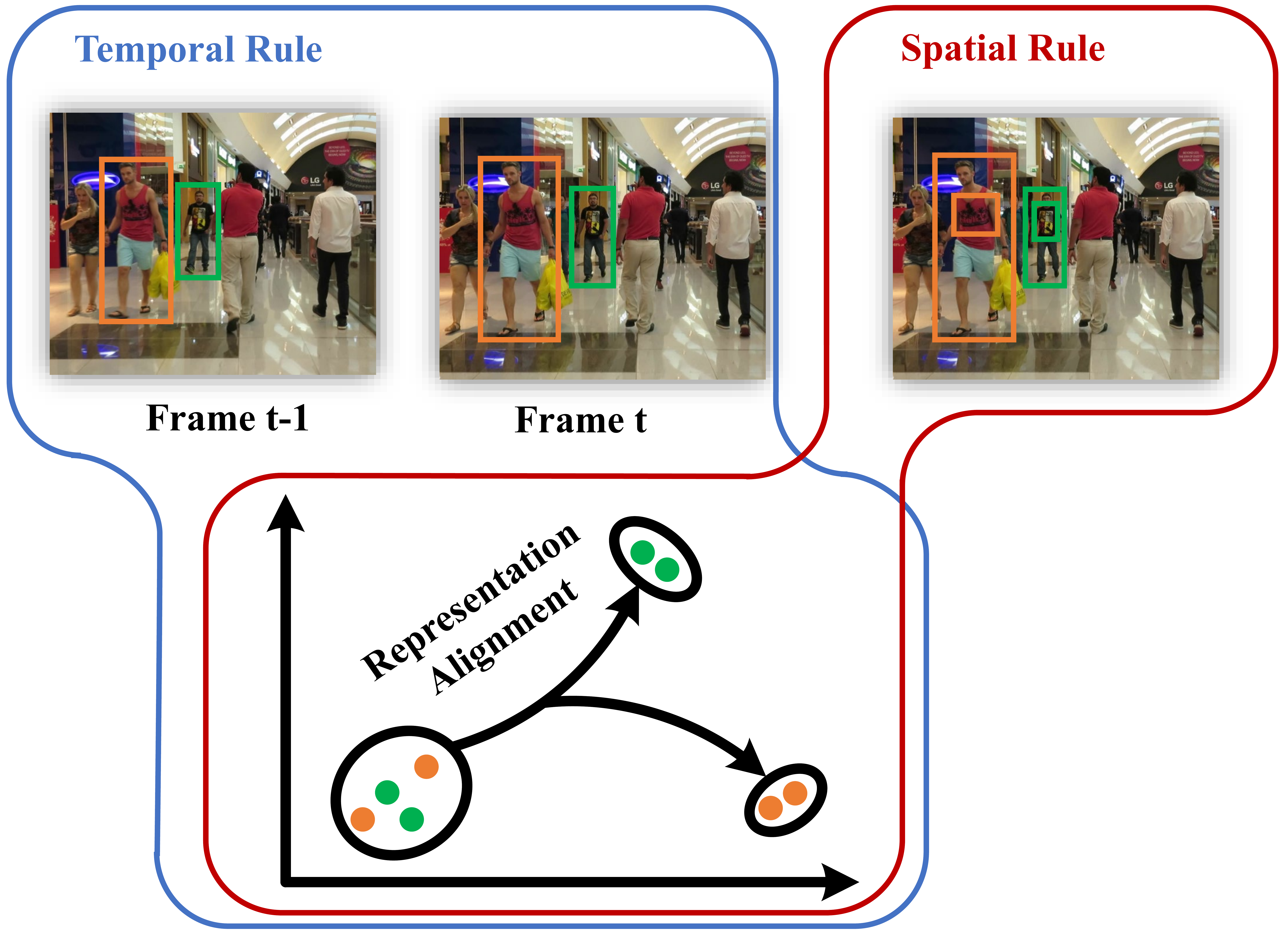}
    \caption{Demonstration of representation alignment rules: Temporal rule reduces gap between consecutive frame's target representations, while spatial rule unites representations of the same object.}
    \label{fig:str_alignment}
\end{figure}

To achieve high-performance tracking results, a diverse array of algorithms and models have been proposed, incorporating spatial and temporal clues. These include methodologies like the Kalman filter \cite{welch1995introduction}, optical flow \cite{baker2004lucas}, memory buffers \cite{cai2022memot, yu2022towards}, Long Short-Term Memory (LSTM) networks \cite{jiang2019graph}, graph-based approaches \cite{wang2021joint,jiang2019graph}, and transformer models \cite{sun2020transtrack,zeng2022motr,dai2022joint}. These algorithms and models aim to leverage spatial and temporal relationships through deep learning frameworks or manually devised association rules. While deep learning frameworks can yield remarkable tracking performance, their association components require meticulous design and can sometimes become intertwined with the foundational architecture. Conversely, rule-based algorithms offer greater flexibility and interpretability, but they lag in performance when faced with objects exhibiting irregular motion patterns.

Is there a simple decoupled module that can effectively model spatio-temporal relationships in principle to suit general tracking scenarios while maintaining excellent tracking performance? To answer this question, we first identify two straightforward yet highly effective rules that can be applied to complex object motions. These two rules are then formalized as contrastive regularization terms for training a lightweight module that doesn't rely on object detection. This detector-free module facilitates the provision of temporally and spatially aligned features, aiding in the improvement of data association.

The two proposed rules provide a broader and contrastive perspective on the alignment of representations, summarizing the spatial and temporal relationships among targets as:
\begin{enumerate}
    \item Representations of same target in consecutive frames should be brought closer, while  representations of different targets should be pushed farther apart. 
    \item  Representations of regions originating from the same target should be brought closer, whereas they should be pushed apart otherwise.
    % Representations of regions that come from same target should be pulled closer to each other, but pushed further otherwise.
\end{enumerate}
The first rule ensures that same objects in consecutive frames are aligned to improve consistency over time. This is based on the idea that the appearance or position of an object doesn't change much between two successive frames.
The second rule focuses on aligning object parts across different regions to enhance spatial consistency. This is guided by the principle that parts of the same category tend to have smaller differences between parts of different categories. When considering the distance between representations of different regions within the same object, it's akin to measuring differences within a category (intra-class difference). On the other hand, the distance between regions from different objects represents differences between categories (inter-class difference). As intra-class differences are typically smaller, regions from the same object should be brought closer in representation space, while regions from different objects should be pushed further apart. Please note that these two rules are not applicable to all scenarios. In certain specific situations, such as tracking rapidly moving objects, the first rule is no longer applicable. Similarly, in cases involving tracking stage actors dressed similarly, the second rule ceases to be applicable. The above two rules are depicted in Figure \ref{fig:str_alignment}.

We apply these two rules as contrastive regularization terms during the training of a module called \textit{Representation Alignment Module} (RAM). The RAM is a versatile component due to its lack of dependency on detectors, allowing it to be seamlessly integrated into any tracker that follows the tracking-by-detection paradigm. It takes the detector outputs as inputs, enhances the features, and generates features that are aligned either spatially or temporally, which are then used for subsequent association steps. The RAM's efficiency lies in its simplicity, as it only requires a single-layer transformer for encoding the aligned features. Additionally, the training overhead and memory requirements are minimal, as the training solely relies on the target's bounding boxes in the video, rather than using complete video frames. We refer to trackers that incorporate RAMs as \textit{RATrackers}.

The key to contrastive regularization lies in creating proper sets of triplets. Hermans \textit{et al.} \cite{hermans2017defense} confirmed that employing an appropriate triplet generation strategy can unleash the tremendous potential of triplets. Inspired by the successful approach of ByteTrack \cite{zhang2022bytetrack}, which employs bounding boxes to achieve state-of-the-art performance, we also utilize bounding boxes as the primary clue for creating triplets.  We reformulate the task of creating these sets as a problem of target association based on bounding boxes. In this setup, the target that corresponds to the anchor target is treated as the positive sample, while the ones that don't match are treated as negative samples. This target association problem has been extensively addressed in existing literature and resolved using conventional optimization techniques \cite{luo2021multiple}.

The contrastive regularization originates from the matching relationship between bounding boxes. Why can the new features improve the performance of data association compared to the original features? We explain this by looking at how the RAM training process resists noise. Due to limitations in traditional optimization methods like the greedy bipartite assignment algorithm \cite{breitenstein2009robust} or the Hungarian algorithm \cite{xing2009multi}, the solutions they provide for the target association problem are occasionally not optimal. This results in mismatched bounding boxes. Triplets made from these mismatches act as noisy samples and can harm the training of RAM. However, because RAM is trained using all triplets, it learns to disregard the noisy ones and produce better features. RAM's ability to filter out noise enhances the quality of aligned features compared to its baseline. To achieve optimal performance, we consider aligned features as complements to the original features. During the association phase, we integrate these aligned features by calculating a weighted sum of affinity matrices. 

The latest works such as QDTrack \cite{pang2021quasi} and MTrack \cite{yu2022towards} also use contrastive regularization for improving association. QDTrack \cite{pang2021quasi} adopts quasi-dense human features for conducting contrast learning while MTrack \cite{yu2022towards} aggregates the whole historical trajectory features. However, it is more likely to include noisy triplets when more candidates are employed, no matter in spatial view as QDTrack \cite{pang2021quasi} did or in temporal view as MTrack \cite{yu2022towards} did. In comparison, we conduct contrast learning on sparse and clean spatial and temporal triplets so as to learn more reliable contrastive regularization. 

Our RAMs demonstrated effectiveness in MOT dataset experiments and minimally impacted the speed of backbone trackers. Additionally, we evaluated RATracker's performance by training it with triplets from detected bounding boxes instead of annotated ones. The results indicated that in unsupervised scenarios, our method only marginally reduced the tracking performance gain, suggesting its capability to enhance tracker performance even without supervision.

The contributions of this paper are in three folds:
\begin{itemize}
    \item Two simple yet effective rules based on representation alignment have been explored for characterizing the spatial and temporal consistency of targets in MOT. They can be formulated as contrastive regularization terms for training RAMs.
    \item A novel, detector-free and lightweight module has been introduced for data association. This module efficiently generates spatially and/or temporally aligned features, seamlessly adaptable across multiple MOT tasks without substantial additional training or memory requirements.
    \item The results from experiments on MOT datasets have confirmed that our proposed rules and RAMs effectively improve the performance of different trackers.
\end{itemize}

\section{Related Work} \label{sec:rel-work}
As our method focus on improving the performance in association stage using contrastive learning method, in what follows we elaborate on most related works of data association and contrastive learning.

\subsection{Data Association}
Data association plays a pivotal role in the field of tracking. The conventional approach to accomplish data association involves affinity computation and bipartite graph matching, as established by Munkres in his work \cite{munkres1957algorithms}.

During the affinity computation phase, three key factors are typically taken into account for linking trajectories and detections: motion \cite{bewley2016simple}, bounding box information \cite{zhang2022bytetrack}, and appearance characteristics \cite{zhang2021fairmot, wang2020towards}. The concept of motion as a clue for association was initially introduced by the SORT algorithm \cite{bewley2016simple}. SORT utilized the Kalman Filter \cite{welch1995introduction} to predict motion in the subsequent frame. Additionally, to capture complex and irregular motions, optical flow was integrated by Xiao et al. \cite{xiao2018simple}. To address challenges posed by substantial camera or object movements, various deep learning-based techniques \cite{sun2020transtrack, wu2021track, zhou2020tracking} were developed.To address more complex scenarios involving nonlinear motion and target occlusion, OC-SORT \cite{cao2023observation} use object observations to compute a virtual trajectory over the occlusion period to fix the error accumulation of filter parameters during the occlusion period. MotionTrack \cite{qin2023motiontrack} utilizes the displacement of targets in the previous frame and employs attention mechanisms to explore the relationships of motion between targets. Bounding box information was employed in the affinity computation process by SORT \cite{bewley2016simple}. ByteTrack \cite{zhang2022bytetrack} proposed a two-stage matching strategy exclusively based on bounding boxes to enhance association performance. Appearance-based clues were favored in the DeepSORT algorithm \cite{wojke2017simple}. This approach utilized a Re-identification (Re-ID) model to extract appearance features and employed the cosine similarity metric for affinity computation. A recent advancement, TransMOT \cite{chu2023transmot}, harnessed a graph transformer to enhance Re-ID features and attain an improved affinity matrix. Some other notable methods like JDE \cite{wang2020towards}, FairMOT \cite{zhang2021fairmot}, and CSTrack \cite{liang2022rethinking} achieved enhanced association results by utilizing appearance features and bounding boxes in separate association stages. 
STRN \cite{xu2019spatial}  leverage spatio-temporal relationships to enhance the original Re-ID features, aiming to maximize the dissimilarity between each target's features.
However, these techniques focused on only one type of clue within each stage.

In the stage of bipartite graph matching, the matching is determined using either the greedy bipartite assignment algorithm as described in Breitenstein et al.'s work \cite{breitenstein2009robust} or the optimal Hungarian algorithm as outlined in Xing et al.'s work \cite{xing2009multi}.

A novel tracking-by-regression approach has been introduced in recent studies, including methods like CenterTrack \cite{zhou2020tracking}, Chained-Tracker \cite{peng2020chained}, TrackFormer \cite{meinhardt2022trackformer}, MOTR \cite{zeng2022motr}, MOTRv2\cite{zhang2023motrv2} and others. In this approach, instead of explicitly associating the current matched bounding boxes with previous trajectories, the bounding boxes of the current frame are directly predicted based on regression, effectively accomplishing the data association implicitly.

\subsection{Contrastive Learning}
Due to its remarkable accomplishments in self-supervised representation learning, contrastive learning has gained widespread adoption across various domains, including classification and action recognition. Prominent examples include the works by He et al. \cite{he2020momentum}, Henaff et al. \cite{henaff2020data}, Tian et al. \cite{tian2020contrastive}, and Wu et al. \cite{wu2018unsupervised}. Furthermore, contrastive learning has recently found application in the field of MOT. The pioneering work of QDTrack \cite{pang2021quasi} introduced contrastive learning to MOT, enhancing appearance features through quasi-dense similarity learning. Subsequently, MTrack \cite{yu2022towards} elevated trajectory representation quality by incorporating complete historical trajectory information and engaging in multi-view trajectory contrastive learning.

Despite these impressive achievements in tracking performance, these methods are susceptible to incorporating more instances of noisy triplets. Additionally, their effectiveness hinges on the availability of annotated matching relationships for facilitating contrastive learning.

\section{Method} \label{sec:method}

\begin{figure*}[tb!]
\centering\includegraphics[width=1.99\columnwidth]{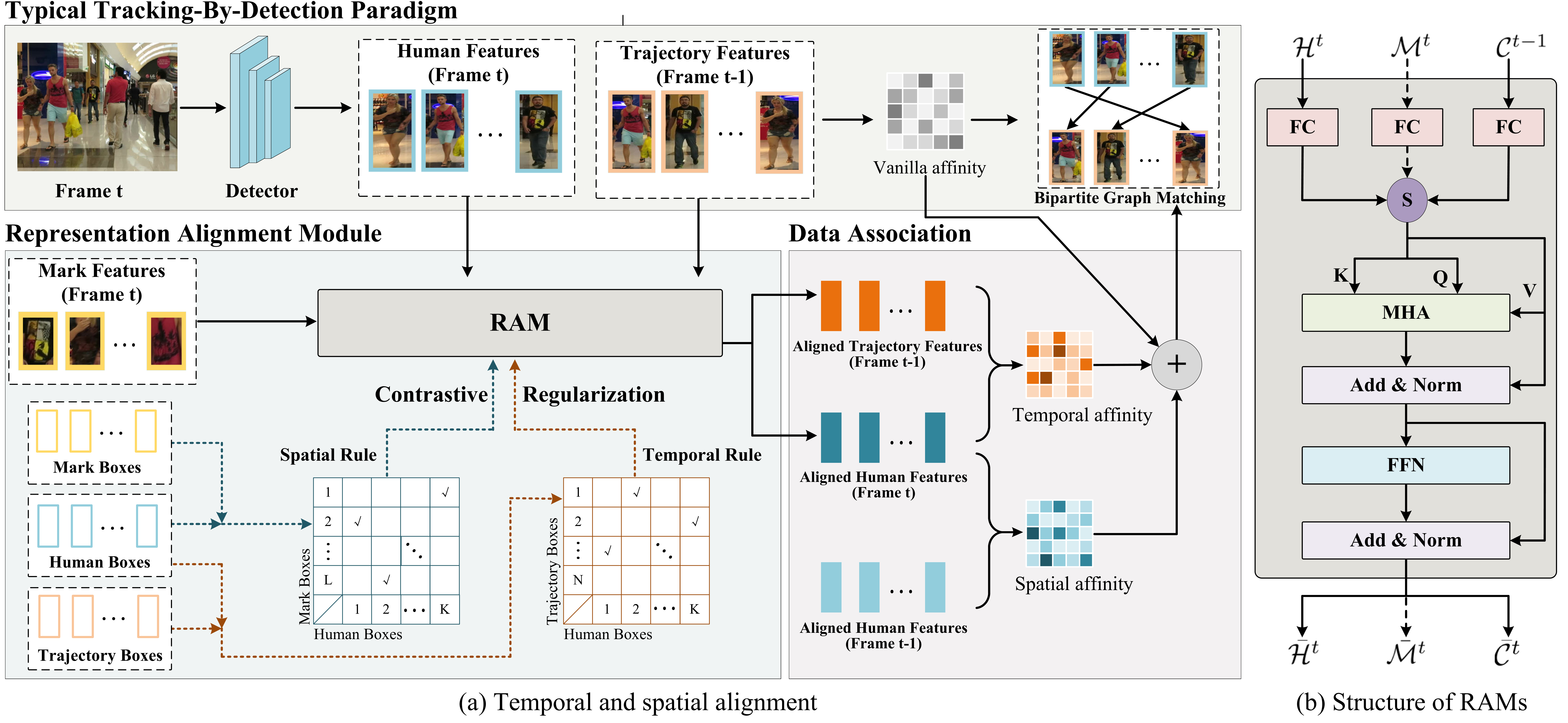}
\caption{The general process of the RATracker and diagram of contrastive regularization derived from representation alignment rules. (a) Diagram of contrastive regularization terms guided by alignment rules, operator \textcircled{+} means weighted sum. (b) Structure of RAMs, operator \textcircled{s} means sequence stack, letters with overbar represent aligned features.  }\label{fig:framework}
\end{figure*}

In this section, we first briefly introduce the overall architecture of RATrackers, then elaborate on the structure of RAMs that incorporate different representation alignment rules, and finally introduce the contrastive regularization for training RAMs.  

\subsection{Overview}
The pipeline of RATrackers follows the tracking-by-detection paradigm. The detector and associator are the same as the backbone tracker, and the only difference lies in the RAM that assists associator, as shown in Figure \ref{fig:framework}. The backbone tracker can be \textit{any} two-stage tracker that conforms to the TBD paradigm, such as FairMOT \cite{zhang2021fairmot}, ByteTrack \cite{zhang2022bytetrack}, CSTrack \cite{liang2022rethinking} and so on. The RAM takes the target features as input, and outputs the aligned features. During the association stage, reliable association is achieved by utilizing the weighted sum of the affinity matrices derived from aligned features and the vanilla features. The rest association steps remain unaltered. 

In what follows, letters with overbar indicates the aligned features. For example, the human features in frame $t$ are characterized by set $\mathcal{H}^t=\{h_i^t\}_{i=1,2,\dots,K}$ and the aligned human features are denoted as set $\bar{\mathcal{H}}^t$.

\subsection{Representation Alignment Module}

The RAM architecture is a simple single-layer transformer encoder that comprises fully connected layers (FCs), a multi-head attention layer (MHA), and a feed-forward network (FFN) as shown in Figure \ref{fig:framework}(b). The FCs transform inputs into higher-dimensional features, the MHA conducts self-attention on these features, and the FFN produces the resultant aligned features. According to rules used for contrastive regularization, RAMs can be divided into temporal RAM, spatial RAM, and spatial-temporal RAM. 

\textbf{Temporal RAM:} 
The Temporal RAM (TRAM) utilizes two fully connected layers to embed inputs, incorporating current human features $\mathcal{H}^t$ and previous trajectory features  $\mathcal{C}^{t-1}$ to generate temporally aligned features $\{\bar{\mathcal{H}}^t, \bar{\mathcal{C}}^t\}$ as outputs. The final affinity matrix $A_T$ for bipartite graph matching is calculated by taking a weighted sum of two affinity matrices. One of these matrices is derived from the original features, while the other comes from temporally aligned features. Given coefficient $\alpha_T \in (0,1)$, the final affinity matrix $A_T$ is computed as 
\begin{equation}\label{eq:temporal_affinity}
A_T=\alpha_T S(\mathcal{H}^t, \mathcal{C}^{t-1}) + (1-\alpha_T) S(\bar{\mathcal{H}}^t, \bar{\mathcal{C}}^t),
\end{equation}
where $S(\cdot, \cdot)$ denotes the similarity function that computes the similarities of two sets. Note that similarity function varies with respect to the type of input features. For example, when inputs are coordinates of bounding boxes, the intersection over union (IoU) metric is used. When inputs are encoded features, the clipped cosine distance over $\mathcal{L}_2$-normed features is preferred.
  
\textbf{Spatial RAM:} 
The spatial RAM (SRAM) operates by taking human features $\mathcal{H}^t$ and mark features $\mathcal{M}^t$ as inputs, producing spatially aligned features $\{\bar{\mathcal{H}}^t, \bar{\mathcal{M}}^t\}$ as outputs. Owing to variations in individual clothing, there exists a distinct region on each person's body that carries identifying information. The mark box serves the purpose of isolating this unique information, yielding characteristic features for establishing associations. They can be generated either by detecting specially designed marks or by following predefined guidelines, such as enclosing 60\% of the area around the center of the detection box. As the aligned mark features $\bar{\mathcal{M}}^t$ are intricately linked to the methodology used for generating the original mark boxes $\mathcal{M}^t$, the aligned human features $\bar{\mathcal{H}}^t$ are more favorable for computing the affinity matrix, a key component in establishing associations. The final affinity matrix $A_S$ for bipartite graph matching is obtained by taking a weighted sum of the original affinity matrix and the affinity matrix derived from spatially aligned features. Given the coefficient $\alpha_S \in (0,1)$, it is computed as 
\begin{equation}\label{eq:spatial_affinity}
A_S=\alpha_S S(\mathcal{H}^t, \mathcal{C}^{t-1}) + (1-\alpha_S) S(\bar{\mathcal{H}}^t, \bar{\mathcal{H}}^{t-1}).
\end{equation}

\textbf{Spatial-Temporal RAM:} The spatial-temporal RAM (STRAM) takes human features $\mathcal{H}^t$, mark features $\mathcal{M}^t$ and previous trajectory features $\mathcal{C}^{t-1}$ as inputs, and outputs the spatially and temporally aligned features $\{\bar{\mathcal{H}}^t, \bar{\mathcal{M}}^t,\bar{\mathcal{C}}^t\}$. The final affinity matrix $A_{ST}$ for bipartite graph matching is computed as the weighted sum of the spatial affinity matrix $A_S$ and temporal affinity matrix $A_T$ with given coefficient $\lambda \in (0,1)$ as
\begin{equation}
A_{ST}=\lambda A_S + (1-\lambda) A_T.
\end{equation}
Note that the purpose of STRAM is to concurrently incorporate spatial and temporal regularization in association. There are multiple approaches to achieve this goal. We opted for a straightforward yet efficient method, which involves a weighted summation of SRAM and TRAM. It's worth considering that employing more intricate fusion techniques might lead to enhanced outcomes.

\subsection{Contrastive Regularization for Training}
In this section, we elaborate on the triplet generation and contrastive regularization for training RAMs under the guidance of the representation alignment rules.

\textbf{Temporal Rule:} 
The primary challenge in implementing the temporal rule is to establish consistent correspondences between the same targets across consecutive frames. This task involves solving an association problem, which can be addressed through affinity computation and bipartite graph matching. The entities to be associated are the detected or annotated human bounding boxes in the current frame and the previous frame. To measure their similarity, the Intersection over Union (IoU) metric is utilized.

To create triplets for training, the human boxes in the previous frame are treated as anchors. For a given anchor and its corresponding counterpart in the current frame, if their IoU surpasses a predefined threshold $\epsilon_{iou}$, the matching outcome is considered reliable, and the counterpart is designated as the positive sample. All other human boxes in the current frame are treated as negative samples in this scenario. When the matching confidence is lower, the anchor itself is treated as the positive sample, and the remaining human boxes in the current frame serve as negative samples.

Furthermore, it is also possible to use human boxes in the current frame as anchors. The process of generating triplets follows a similar procedure as described above. In this case, we employ the InfoNCE (Noise Contrastive Estimation) loss \cite{he2020momentum} to establish contrast between samples. The contrastive loss derived from the initial set of triplets constitutes the forward temporal loss, while the loss from the latter set of triplets forms the backward temporal loss. The overall temporal loss is the combination of both forward and backward temporal losses and can be calculated as outlined below.

\vspace{-4mm}
\begin{equation}
\resizebox{\columnwidth}{!}{$\begin{aligned}
\mathcal{L}_{T}=-\sum_{v_a\in \bar{\mathcal{H}}^t}\log\dfrac{\exp{(v_a\cdot v^{t-1}_p/\tau)}}
{\exp{(v_a\cdot v^{t-1}_p/\tau)}+\sum\limits_{v_n\in \bar{\mathcal{H}}_n^{t-1}}\exp{(v_a\cdot v_n/\tau)}}\\
-\sum_{v_a\in \bar{\mathcal{H}}^{t-1}}\log\dfrac{\exp{(v_a\cdot v^t_p/\tau)}}
{\exp{(v_a\cdot v^t_p/\tau)} +\sum\limits_{v_n\in \bar{\mathcal{H}}^t_n}\exp{(v_a\cdot v_n/\tau)}},
\end{aligned}$}
\end{equation}
where $\tau$ is the temperature hyper-parameter, $v_a, v_p, v_n$ are anchor feature, positive feature and negative feature respectively. The set $\bar{\mathcal{H}}_n$ is the negative feature set with respect to the anchor feature $v_a$. 

\textbf{Spatial Rule:} 
The primary challenge in implementing spatial rule is to establish connections between targets that belong to the same object within a single frame. This challenge can be addressed using affinity computation and bipartite graph matching. In this context, the targets to be matched are represented by human boxes and mark boxes.

However, the conventional Intersection over Union (IoU) metric is not suitable for measuring similarity in this scenario. IoU struggles to differentiate between a mark and two occluded human bodies due to its heavy reliance on the size of the bodies. This can lead to erroneous matches, where the IoU of a foreground mark and a background human with a smaller body size might exceed the actual ground truth, resulting in inaccurate matches. To mitigate this issue, we have observed that using the size of the mark as a basis for normalization of the intersection provides greater reliability, as it remains invariant to human occlusion. Consequently, we've introduced a more robust metric called \textit{Intersection Rate} (IR) for computing the affinity matrix. The IR metric quantifies the intersection of the mark box and the human box relative to the mark box itself.

To generate triplets for spatial contrast, we adopt the following approach: When the mark box serves as the anchor, a human box that shares an IR value surpassing a certain threshold $\epsilon_{ir}$ is designated as the positive sample, while all other human boxes become negative samples. Alternatively, if the IR threshold isn't met, the mark box itself is designated as the positive sample, and all human boxes are classified as negative samples. The same process is mirrored when the human box is utilized as the anchor.

The comprehensive spatial contrastive loss can be calculated in the ensuing manner:
\begin{equation}
\resizebox{\columnwidth}{!}{$\begin{aligned}
\mathcal{L}_{S}=-\sum_{h_a\in \bar{\mathcal{H}}^t}\log\dfrac{\exp{(h_a\cdot m_p/\tau)}}
{\exp{(h_a\cdot m_p/\tau)}+\sum\limits_{m_n\in \bar{\mathcal{M}}_n^t}\exp{(h_a\cdot m_n/\tau)}}\\
-\sum_{m_a\in \bar{\mathcal{M}}^t}\log\dfrac{\exp{(m_a\cdot h_p/\tau)}}
{\exp{(m_a\cdot h_p/\tau)} +\sum\limits_{h_n\in \bar{\mathcal{H}}^t_n}\exp{(m_a\cdot h_n/\tau)}},
\end{aligned}$}
\end{equation}
where $h_a$ is the anchor human feature, $m_n, m_p$ are the negative and positive mark features respectively, $\bar{\mathcal{M}}_n^t$ is the set of negative mark features w.r.t. the anchor feature $h_a$. Similarly, $m_a$ is the anchor mark feature, $h_n, h_p$ are the negative and positive human features respectively, $\bar{\mathcal{H}}_n^t$ is the set of negative human features w.r.t. the anchor feature $m_a$.

\textbf{Spatial-Temporal Rule:} The implementation of spatial-temporal rule simply replicates the implementation of spatial rule and temporal rule simultaneously. The overall contrastive loss can be computed as 
\begin{equation}
    \mathcal{L}_{ST} = \mathcal{L}_{S} + \mathcal{L}_{T}.
\end{equation}

\section{Experiments} \label{sec:eval}

We conduct extensive experiments over three publicly accessible datasets including MOT17 \cite{milan2016mot16}, MOT20 \cite{dendorfer2020mot20} and BDD100K \cite{yu2020bdd100k}. We used the ID score \cite{ristani2016performance} and CLEAR MOT metrics \cite{bernardin2008evaluating} to evaluate the performance of the proposed method. Throughout the experiments, we generated the mark boxes by boxing out 60\% of the area around the center of the detection box. The input features can either be bounding boxes characterized by corner coordinates and box-size as $(x, y, h, w)$, or the Re-ID features extracted from some pretrained modules like fastReID \cite{he2020fastreid}. Unless specified otherwise, our input features are assumed to be bounding boxes.

\textbf{Training Details:} 
The experiments were carried out using PyTorch and an NVIDIA GeForce RTX 2080 Ti GPU. The training process involved running for 50 epochs with a batch size of 5. The chosen optimizer was AdamW \cite{loshchilov2017decoupled}, initialized with a learning rate of $2\times 10^{-3}$, which decreased by a factor of 10 every 10 epochs. To accommodate input sequences of varying lengths, a strategy inspired by DETR \cite{carion2020end} was applied. Input sequences were standardized to a fixed length of 110 for MOT17, 260 for MOT20 and 100 for BDD100K. This was achieved by appending invalid bounding boxes with all zero coordinates. Notably, these added boxes were disregarded during loss calculations. The output dimension of the fully connected layers was configured to be 128, aligning with the approach.

\textbf{Hyperparameters:} 
During the course of the experiments, parameters $\lambda=0.5, \tau=0.1, \epsilon_{ir}=0$ were consistently configured. The selection of parameters $\alpha_S$, $\alpha_T$, and $\epsilon_{iou}$, however, varied based on the particular experiment being conducted. In cases where the ByteTrack \cite{zhang2022bytetrack} backbone tracker employed two association stages, parameters $\alpha_S=\alpha_T=0.2$, $\epsilon_{iou}=0.9$ were chosen for the initial stage, and parameters $\alpha_S=\alpha_T=0.3$, $\epsilon_{iou}=0.5$ were employed for the subsequent stage. In instances where only one association stage was utilized such as TransTrack\cite{sun2020transtrack}, parameters $\alpha_S=\alpha_T=0.3$, $\epsilon_{iou}=0.9$ were employed.

\subsection{Effectiveness of RAMs}

\subsubsection{On Different Trackers}
\begin{table}[tb!]
\centering
\renewcommand{\arraystretch}{1.2}
% \scalebox{0.8}{
\resizebox{1\columnwidth}{!}{
\begin{tabular}{@{} *{1}l| *{3}l @{}}
\toprule
\textbf{Method} & \textbf{IDF1} $\uparrow$ & \textbf{MOTA} $\uparrow$ & \textbf{IDS} $\downarrow$ \\
% \cmidrule{1-4}
% FairMOT \cite{zhang2021fairmot} &  72.81 & 69.06 & 299  \\
% FairMOT+TRAM &  74.44(\textbf{+1.63}) & 69.37(\textbf{+0.31}) & 272(\textcolor{blue}{\textbf{-23}})  \\
% FairMOT+SRAM &  74.02(\textbf{+1.21}) & 69.13(\textbf{+0.07}) & 290(\textbf{-9})  \\
% FairMOT+STRAM &  74.67(\textcolor{blue}{\textbf{+1.86}}) & 69.38(\textcolor{blue}{\textbf{+0.32}}) & 289(\textbf{-10})  \\
\cmidrule{1-4}
JDE \cite{wang2020towards} & 63.59 & 59.98 & 473  \\
JDE+TRAM  &  67.30(\textcolor{blue}{\textbf{+3.71}}) & 60.31(\textbf{+0.33}) & 383(\textbf{-80})  \\
JDE+SRAM  &  66.75(\textbf{+3.16}) & 60.29(\textbf{+0.31}) & 372(\textcolor{blue}{\textbf{-101}})  \\
JDE+STRAM  &  67.20(\textbf{+3.61}) & 60.47(\textcolor{blue}{\textbf{+0.49}}) & 374(\textbf{-99})  \\
\cmidrule{1-4}
CSTrack \cite{liang2022rethinking} & 71.82 & 67.96 & 340  \\
CSTrack+TRAM  &  73.56(\textbf{+1.74}) & 68.52(\textbf{+0.56}) & 260(\textcolor{blue}{\textbf{-50}})  \\
CSTrack+SRAM  &  72.93(\textbf{+1.11}) & 68.46(\textbf{+0.5}) & 304(\textbf{-36})  \\
CSTrack+STRAM  &  73.70(\textcolor{blue}{\textbf{+1.88}}) & 68.63(\textcolor{blue}{\textbf{+0.67}}) & 291(\textbf{-49})  \\
\cmidrule{1-4}
TransTrack \cite{sun2020transtrack} & 68.60 & 67.66 & 254  \\
TransTrack+TRAM &  71.64(\textcolor{blue}{\textbf{+3.04}}) & 67.86(\textbf{+0.2}) & 245(\textbf{-9})  \\
TransTrack+SRAM &  69.71(\textbf{+1.11}) & 67.85(\textbf{+0.19}) & 250(\textbf{-4})  \\
TransTrack+STRAM &  71.14(\textbf{+2.54}) & 67.98(\textcolor{blue}{\textbf{+0.32}}) & 238(\textcolor{blue}{\textbf{-16}})  \\
\cmidrule{1-4}
ByteTrack \cite{zhang2022bytetrack} & 79.07 & 76.49 & 165  \\
ByteTrack+TRAM & 79.92(\textbf{+0.85}) & 76.82(\textbf{+0.33}) & 145(\textbf{-18})  \\
ByteTrack+SRAM & 79.90(\textbf{+0.83}) & 76.82(\textbf{+0.33}) & 139(\textcolor{blue}{\textbf{-26}})  \\
ByteTrack+STRAM & 80.87(\textcolor{blue}{\textbf{+1.8}}) & 76.90(\textcolor{blue}{\textbf{+0.41}}) & 155(\textbf{-10})  \\
\cmidrule{1-4}
OC-SORT \cite{cao2023observation} & 77.85 & 74.12 & 195  \\
OC-SORT+TRAM & 78.09(\textbf{+0.24}) & 74.35(\textbf{+0.23}) & 169(\textbf{-23})  \\
OC-SORT+SRAM & 78.07(\textbf{+0.22}) & 74.21(\textbf{+0.09}) & 192(\textbf{-3})  \\
OC-SORT+STRAM & 78.72(\textcolor{blue}{\textbf{+0.87}}) & 74.38(\textcolor{blue}{\textbf{+0.26}}) & 164(\textcolor{blue}{\textbf{-31}})  \\
\bottomrule
\end{tabular}}
\vspace{2mm}
\caption{Results of applying RAMs to five popular trackers on the MOT17 validation set. $\uparrow$ means higher is better, $\downarrow$ means lower is better}
\label{tab:flexibility_validation}
\end{table}

This evaluation encompasses five state-of-the-art trackers employing RAMs: JDE \cite{wang2020towards}, CSTrack \cite{liang2022rethinking}, TransTrack \cite{sun2020transtrack}, ByteTrack \cite{zhang2022bytetrack} and OC-SORT \cite{cao2023observation}. All of these trackers adhere to the tracking-by-detection paradigm and involve the computation of affinity matrices for the purpose of association. The experiment was carried out on the MOT17 validation dataset.

Table \ref{tab:flexibility_validation} illustrates that integrating RAMs consistently improves crucial performance metrics such as MOTA, IDF1, and IDS across various trackers. Among the trackers studied, CSTrack \cite{liang2022rethinking} and JDE \cite{wang2020towards} utilize CNNs for feature extraction in association, while TransTrack \cite{sun2020transtrack} uses the Transformer architecture for feature generation. ByteTrack \cite{zhang2022bytetrack} directly employs bounding boxes, and OC-SORT \cite{cao2023observation} refines them through a motion prediction model. These findings emphasize how RAMs enhance the performance of different trackers, irrespective of their specific backbone frameworks, showcasing their versatile applicability. 

\begin{figure}[tb]
    \centering
    \includegraphics[width=1\columnwidth]{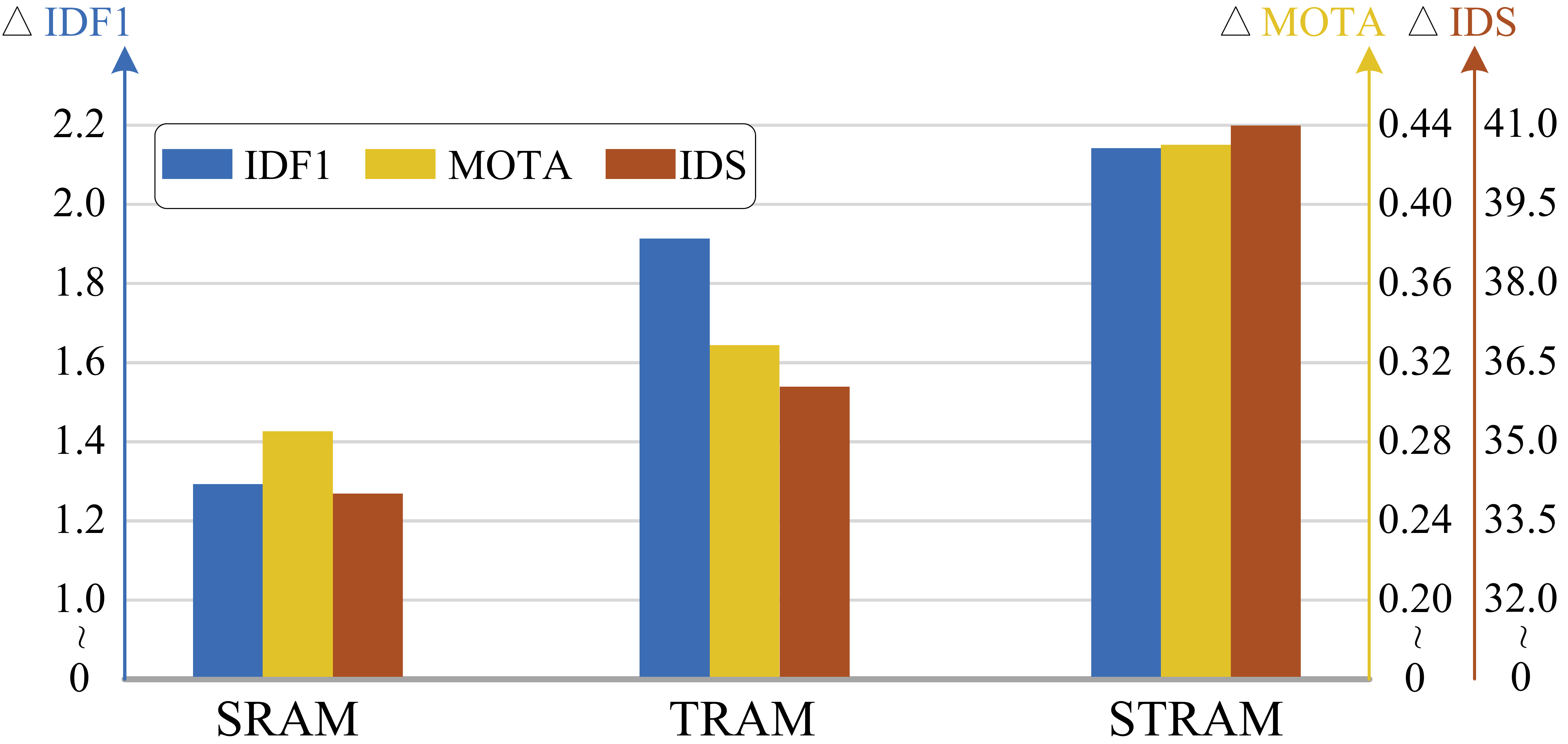}
    \caption{The average performance of RAMs on various trackers in Table \ref{tab:flexibility_validation}.}
    \label{fig:performance_gains}
\end{figure}

Figure \ref{fig:performance_gains} provides the average performance of RAMs across various trackers. While individual metrics for TRAM might surpass STRAM in specific backbone trackers as indicated in Table \ref{tab:flexibility_validation}, overall, STRAM consistently outperforms both SRAM and TRAM. This indicates that, in general, considering the performance of spatio-temporal alignment regularization is superior to solely focusing on single-branch regularization.

\subsubsection{On Different Datasets}

\begin{table*} [tb!]
\renewcommand{\arraystretch}{1.2}
\centering 
\scalebox{0.75}{
\begin{tabular}{@{} *{1}l|*{3}l|*{3}l|*{3}l|*{3}l @{}}
  
\toprule
& \multicolumn{3}{c|}{MOT17-val} & \multicolumn{3}{c|}{MOT20-val} & \multicolumn{3}{c|}{BDD100K-val}  & \multicolumn{3}{c}{Average Performance} \\

\cmidrule{1-13}
 & IDF1$\uparrow$  & MOTA$\uparrow$  & IDS$\downarrow$  & IDF1$\uparrow$  & MOTA$\uparrow$  & IDS$\downarrow$ &
 IDF1$\uparrow$  & MOTA$\uparrow$  & IDS$\downarrow$ & IDF1$\uparrow$  & MOTA$\uparrow$  & IDS$\downarrow$ \\

\cmidrule{1-13}
 Baseline  & 75.56   & 79.85  & 495  & 81.62 & 77.90 & 913 & 54.95   & 45.11    & 32963  & 70.71 & 67.62 & 11457\\
  Baseline+TRAM  & 76.34   & 80.51  & \textbf{478}  & 81.73 & \textbf{78.08} & 898 & 55.54   & 45.41    & 31372 & 71.20 & 68.00 & 10916 \\
   Baseline+SRAM  & 75.94   & 80.50  & 480  & 81.74 & 77.95 & 897 & 55.13   & 45.16    & 32104 & 70.94 & 67.87 & 11160\\
    Baseline+STRAM     & \textbf{77.14}   & \textbf{81.14}  & 479  & \textbf{81.86} & 77.92 & \textbf{896} & \textbf{55.90}   & \textbf{45.55}    & \textbf{30567} & \textbf{71.63} & \textbf{68.20} & \textbf{10647} \\
\bottomrule
\end{tabular}
}
\caption{The performance of RAMs on multiple datasets. The best results are marked in \textbf{bold}}
\label{tab:multi-dataset-validation}
\end{table*}

\begin{table*} [tb!]
\renewcommand{\arraystretch}{1.2}
\centering 
\scalebox{0.75}{
\begin{tabular}{@{}l|*{1}l|*{9}l @{}}
\hline

\toprule
\textbf{Benchmark}   &\textbf{Method}   &\textbf{MOTA}$\uparrow$  &\textbf{IDF1}$\uparrow$  &\textbf{HOTA}$\uparrow$  &\textbf{AssA}$\uparrow$  &\textbf{MT}$\uparrow$   &\textbf{ML}$\downarrow$  &\textbf{FP}$\downarrow$     &\textbf{FN}$\downarrow$    &\textbf{IDS}$\downarrow$  \\
\cmidrule{1-11}
\multirow{17}{*}{MOT17}
& RelationTrack \cite{yu2022relationtrack}  & 73.8   & 74.7  & 61.0  & 61.5 & 41.7 & 23.2 & 27999   & 118623    & \underline{1374} \\
& CenterTrack \cite{zhou2020tracking} & 67.8 & 64.7 & 52.2 & - & 34.6 & 24.6 & \underline{18489} & 160332 & 3039     \\
& TraDeS \cite{wu2021track} & 69.1 & 63.9 & 52.7 & 50.8 & 36.4 & 21.5 & 20892 & 150060 & 3555   \\
& CorrTracker \cite{wang2021multiple} &  76.5 & 73.6 & 60.7 & 58.9 & 47.6 & 12.7 & 29808 & 99510 & 3369   \\
& CTracker \cite{peng2020chained} &  66.6 & 57.4 & 49.0 & 45.2 & 32.2 & 24.2 & 22284 & 160491 & 5529  \\
& QDTrack \cite{pang2021quasi} &  68.7 & 66.3 & 53.9 & 52.7 & 40.6 & 21.9 & 26589 & 146643 & 3378 \\
& MTrack \cite{yu2022towards} &  72.1 & 73.5 & 60.5 & 60.9 & 49.0 & 16.8 & 53361 & 101844 & 2028  \\
& TransCenter \cite{xu2021transcenter} & 73.2 & 62.2 & 54.5 & 49.7 & 40.8 & 18.5 & 23112 & 123738 & 4614  \\
& MOTR \cite{zeng2022motr} &  78.6 & 75.0 & 62.0 & 60.6 & 50.3 & 13.1 & 23409 & 94797 & 2619 \\
& TransMOT \cite{chu2023transmot} & 76.7 & 75.1 & 61.7 & - & 51.0 & 16.4 & 36231 & 93150 & 2346 \\
& TransTrack \cite{sun2020transtrack} &  75.2 & 63.5 & 54.1 & 47.9 & 55.3 & \textbf{10.2} & 50157 & 86442 & 3603 \\
& TrackFormer \cite{meinhardt2022trackformer} &  74.1 & 68.0 & 57.3 & 54.1 & 47.3 & \underline{10.4} & 34602 & 108777 & 2829 \\
& MeMOT \cite{cai2022memot} &  72.5 & 69.0 & 56.9 & 55.2 & 43.8 & 18.0 & 37221 & 115248 & 2724 \\
& CSTrack \cite{liang2022rethinking} &  74.9 & 72.6 & 59.3 & 57.9 & 41.5 & 17.5 & 23847 & 114303 & 3567 \\
& FairMOT \cite{zhang2021fairmot} &  73.7 & 72.3 & 59.3 & 58.0 & 43.2 & 17.3 & 27507 & 117477 & 3303 \\
& ByteTrack \cite{zhang2022bytetrack} &  80.3 & 77.3 & 63.1 & 62.0 & 53.2 & 14.5 & 25491 & 83721 & 2196 \\
& MOTRv2 \cite{zhang2023motrv2} &  78.6 & 75.0 & 62.0 & 60.6 & - & - & - & - & - \\
& OC-SORT \cite{cao2023observation} &  78.0 & 77.5 & 63.2 & 63.2 & - & - & \textbf{15100} & 108000 & 1950 \\
& MotionTrack \cite{qin2023motiontrack} &  \textbf{81.1} & \textbf{80.1} & \textbf{65.1} & \textbf{65.1} & \underline{55.5} & 16.7 & 23802 & \underline{81660} & \textbf{1140} \\
& \textbf{ByteTrack+STRAM(ours)} &   \underline{81.0} & \underline{79.9} & \underline{64.9} & \underline{64.8} & \textbf{56.2} & 14.4 & 24459 & \textbf{81198} & 1383 \\

\cmidrule{1-11}
% \hline
\multirow{9}{*}{MOT20} 
& TransCener \cite{xu2021transcenter} &  58.5 & 49.6 & 43.5 & 37.0 & 48.6 & 14.9 & 64217 & 146019 & 4695 \\
& RelationTrack \cite{yu2022relationtrack} &  67.2 & 70.5 & 56.5 & 56.4 & 62.2 & 8.9 & 61134 & 104597 & 4243 \\
& MeMOT \cite{cai2022memot} &  63.7 & 66.1 & 54.1 & 55.0 & 57.5 & 14.3 & 47882 & 137983 & 1938 \\
& MTrack \cite{yu2022towards} &  63.5 & 69.2 & 55.3 & 55.7 & 68.8 & \textbf{7.5} & 96123 & \underline{86964} & 6031\\
& TransTrack \cite{sun2020transtrack} &  65.0 & 59.4 & 48.9 & 45.2 & 50.1 & 13.4 & 27191 & 150197 & 3608 \\
& CSTrack \cite{liang2022rethinking} &  66.6 & 68.6 & 54.0 & 54.0 & 50.4 & 15.5 & 25404 & 144358 & 3196  \\
& FairMOT \cite{zhang2021fairmot} &  61.8 & 67.3 & 54.6 & 54.7 & 68.8 & \underline{7.6} & 103440 & 88901 & 5243  \\
& ByteTrack \cite{zhang2022bytetrack} &  77.8 & 75.2 & 61.3 & 59.6 & 69.2 & 9.5 & 26249 & 87594 & 1223 \\
& MOTRv2 \cite{zhang2023motrv2} &  76.2 & 73.1 & 61.0 & 59.3 & - & - & - & - & - \\
& OC-SORT \cite{cao2023observation} &  75.5 & 75.9 & 62.1 & \underline{62.0} & - & - & \textbf{18000} & 108000 & \textbf{913} \\
& MotionTrack \cite{qin2023motiontrack} &  \textbf{78.0} & \underline{76.5} & \underline{62.8} & 61.8 & \textbf{71.3} & 9.5 & 28629 & \textbf{84152} & \underline{1165} \\
& \textbf{ByteTrack+STRAM(ours)} &  \underline{77.9} & \textbf{77.3} & \textbf{63.3} & \textbf{62.8} & \underline{70.3} & 9.6 & \underline{24353} & 88867 & 1309\\
\bottomrule
% \hline
\end{tabular}
}
\caption{Performance comparison with preceding SOTAs on the testing splits of the MOT17 and MOT20 benchmarks under the private detection protocols. The best results are marked in \textbf{bold} and the suboptimal results are annotated with \underline{underline}}
\label{tab:mot_benchmark}
\end{table*}

We validated the effectiveness of RAMs on three datasets: MOT17 \cite{milan2016mot16}, MOT20 \cite{dendorfer2020mot20}, and BDD100K \cite{yu2020bdd100k}. MOT17 and MOT20 are popular pedestrian tracking datasets, with MOT20 having a higher density of pedestrians. BDD100K is a large dataset used for vehicle tracking, comprising 2000 training and testing scenes. To ensure a fair comparison, we introduced a baseline tracking method using YOLOVX \cite{ge2021yolox} for object detection, Kalman filtering for trajectory prediction, and bounding boxes as the association features.

Table \ref{tab:multi-dataset-validation} illustrates that across various datasets, the use of RAMs consistently enhances the performance of the baseline method. Notably, there's a significant improvement in metrics for MOT17 and BDD100K datasets compared to a more marginal enhancement in MOT20. Additionally, TRAM outperforms STRAM in terms of MOTA in MOT20, likely due to MOT20 containing denser objects with higher chances of visual similarity, potentially causing spatial alignment regularization to be less effective. However, considering the overall results, STRAM still outperforms both single-branch TRAM and SRAM approaches.

\subsubsection{On MOT Benchmarks}

We use ByteTrack\cite{zhang2022bytetrack} as the backbone tracker and evaluate the performance of ByteTrack+STRAM in the MOT17 and MOT20 benchmarks using a private detection setup. We train STRAMs separately using the training sets from MOT17 and MOT20. To evaluate performance, we submit the tracking results from the test sets to the official MOT Challenge evaluation platform.

Table \ref{tab:mot_benchmark} demonstrates ByteTrack+STRAM's impressive performance. On the MOT17 dataset, it achieves significant scores of 81.0 MOTA and 79.9 IDF1. Even on the more complex MOT20 benchmark, it maintains strong results with 77.9 MOTA and 77.3 IDF1. Notably, both false positive (FP) and false negative (FN) metrics remain minimal for both MOT17 and MOT20. This indicates that STRAM effectively reduces incorrectly tracked boxes and successfully re-establishes tracking for previously overlooked targets through associations. The exceptional performance is credited to STRAM's integration of spatial and temporal rule-based contrastive regularization terms.

\subsubsection{Computational Complexity}

% \vspace{-2mm}
\begin{table} [h]
\renewcommand{\arraystretch}{1.2}
\centering 
% \scalebox{1.2}{
\resizebox{1\columnwidth}{!}{
\begin{tabular}{@{} *{1}l|*{3}l @{}}
\toprule
\textbf{Method} & \textbf{Params(M)} & \textbf{Flops(G)} & \textbf{FPS} \\
\cmidrule{1-4}
FairMOT  &  16.5542 &  72.932 &  22.5 \\
FairMOT+TRAM &  16.6219 &  72.939 &  21.5 \\
FairMOT+SRAM &  16.6219 &  72.939 &  21.4 \\
FairMOT+STRAM &  16.6226 &  72.942 &  21.1 \\
\cmidrule{1-4}
ByteTrack  & 98.9954  & 793.211  & 25.7  \\
ByteTrack+TRAM & 99.0677  & 793.218  & 25.0  \\
ByteTrack+SRAM & 99.0677  & 793.218  & 25.0  \\
ByteTrack+STRAM & 99.0684  & 793.221  & 23.8  \\
\bottomrule
\end{tabular}}
\caption{Computational complexity results on MOT17 test set}
\label{tab:table_params_flops_fps}
\end{table}

We performed an experiment to assess computational complexity using a single NVIDIA GeForce RTX 3090 Ti GPU. The outcomes are outlined in Table \ref{tab:table_params_flops_fps}. Because the RAM employs only a single-layer transformer and the input sequence length is typically short (equivalent to the number of targets in each frame), the extra parameters and computational load are minimal. This has a negligible impact on the real-time performance of the original tracker.

\subsection{Ablation Study}

\subsubsection{Features for RAMs and Affinity Computation}

The contrastive regularization triplets are exclusively derived from bounding boxes. However, when utilized in RAMs and computing affinity matrices, there exists flexibility in the types of features employed. Our experiments on MOT17 validation set, testing diverse input feature types in RAMs and affinity computation, showcase the resilience of our proposed method across these variations.

\begin{table} [tb!]
\renewcommand{\arraystretch}{1.2}
\centering 
% \scalebox{0.9}{
\resizebox{1\columnwidth}{!}{
\begin{tabular}{@{} *{1}l|*{1}l|*{3}l @{}}
\toprule
\textbf{Type of RAMs} & \textbf{Input Feature} & \textbf{IDF1} $\uparrow$ & \textbf{MOTA} $\uparrow$ & \textbf{IDS} $\downarrow$ \\
\cmidrule{1-5}
Without RAM & -  &  72.81 & 69.06 & 299  \\
\cmidrule{1-5}
\multirow{2}{*}{TRAM}
& Bounding Box &  74.44(\textbf{+1.63}) & 69.37(\textbf{+0.31}) & 272(\textbf{-23})  \\
& Re-ID &  73.93(\textbf{+1.12}) & 69.18(\textbf{+0.12}) & 250(\textbf{-49})  \\
\cmidrule{1-5}
\multirow{2}{*}{SRAM}
& Bounding Box &  74.02(\textbf{+1.21}) & 69.13(\textbf{+0.07}) & 290(\textbf{-9})  \\
& Re-ID &  74.36(\textbf{+1.55}) & 69.21(\textbf{+0.15}) & 267(\textbf{-32})  \\
\cmidrule{1-5}
\multirow{2}{*}{STRAM}
& Bounding Box &  74.67(\textbf{+1.86}) & 69.38(\textbf{+0.32}) & 289(\textbf{-10})  \\
& Re-ID &  73.92(\textbf{+1.11}) & 69.32(\textbf{+0.26}) & 240(\textbf{-59})  \\
\bottomrule
\end{tabular}}% end of scalebox
\caption{Results of RAMs to different input features with FairMOT\cite{zhang2021fairmot} backbone on the MOT17 validation set. The best results are marked in blue}
\label{tab:reid_based_ram}
% \vspace{4mm}
\end{table}

RAMs can intake appearance features such as Re-ID features or bounding boxes. To implement the RAM module with appearance features, we utilized FastReID \cite{he2020fastreid} for extracting Re-ID features. Table \ref{tab:reid_based_ram} demonstrates the performance enhancements achieved by applying Re-ID features or bounding box features to FairMOT with RAMs. Notably, TRAM, SRAM, and STRAM consistently contributed to a stable improvement of at least 1 in IDF1, regardless of the input feature types.

The robustness of our proposed method across diverse association features is delineated in Table \ref{tab:ablation_study_st}. Utilizing ByteTrack\cite{zhang2022bytetrack} as the backbone tracker, we evaluated its tracking performance using various association feature combinations, with detection score thresholds set at $\tau_{high}=0.6$ and $\tau_{low}=0.1$. The results in Table \ref{tab:ablation_study_st}, focusing on two association stages employing the same type of feature, consistently indicate that incorporating STRAM enhances tracking performance, irrespective of whether IoU, Re-ID, or a combination of both is considered. This underscores the adaptability of RAM in improving various association features.

\begin{table} [tb!]
\renewcommand{\arraystretch}{1.2}
\centering 
% \scalebox{0.8}{
\resizebox{1\columnwidth}{!}{
\begin{tabular}{@{} *{1}l|*{3}l @{}}
\toprule
\textbf{Association Features} & \textbf{IDF1} $\uparrow$ & \textbf{MOTA} $\uparrow$ & \textbf{IDS} $\downarrow$ \\
\cmidrule{1-4}
Bounding Box &  79.55 & 77.65 & 333  \\
Bounding Box + STRAM & 81.01(\textbf{+1.46}) & 77.94(\textbf{+0.29}) & 267(\textbf{-66})  \\
\cmidrule{1-4}
Re-ID & 70.43 & 73.27 & 447  \\
Re-ID + STRAM & 77.95(\textbf{+7.52}) & 75.25(\textbf{+1.98}) & 370(\textbf{-77})  \\
\cmidrule{1-4}
Bounding Box + Re-ID  & 79.07 & 77.73 & 223  \\
Bounding Box + Re-ID + STRAM & 81.82(\textbf{+2.75}) & 78.21(\textbf{+0.48}) & 203(\textbf{-20})  \\
\bottomrule
\end{tabular}}
\caption{Results of ByteTrack \cite{zhang2022bytetrack} to various association features on the MOT17 validation set with and without STRAM}
\label{tab:ablation_study_st}
% \vspace{4mm}
\end{table}

\subsubsection{Embedding Dimension}

We conducted an experiment focusing on the output dimension of the FC layers within the STRAM. The backbone tracker used for this experiment was ByteTrack\cite{zhang2022bytetrack}. The results are summarized in Table \ref{tab:table-feature-dim-ablation-study}. We achieved the highest values of 80.87 for IDF1 and 76.90 for MOTA when the output dimension was set to 128. Conversely, the lowest value of 154 for IDS was observed when the output dimension was set to 1024. There was only marginal improvement in tracking performance with increasing dimensions. Consequently, for consistency, we maintained an embedding dimension of 128 throughout the experiments detailed in the paper.

\begin{table} [tb!]
\renewcommand{\arraystretch}{1.2}
\centering 
% \scalebox{1}{
\resizebox{1\columnwidth}{!}{
    \begin{tabular}{@{} *{1}c|*{3}l @{}}
    \toprule
    \textbf{Embedding Dimension} & \textbf{IDF1} $\uparrow$ & \textbf{MOTA} $\uparrow$ & \textbf{IDS} $\downarrow$ \\
    
    \cmidrule{1-4}
    64  & 80.18 & 76.81 & 157  \\
    128  & \textbf{80.87} & \textbf{76.90} & 155  \\
    256 & 80.26 & 76.85 & 162  \\
    512 & 80.67 & 76.63 & 164  \\
    1024 & 80.16 & 76.77 & \textbf{154}  \\
    
    \bottomrule
    \end{tabular}
}% end of scalebox
\caption{Exploring STRAM performance across various FC layer embedding dimensions on MOT17 validation set}
\label{tab:table-feature-dim-ablation-study}
% \vspace{2mm}
\end{table}

\subsubsection{Hyperparameters}

\begin{figure}[tb]
    \centering
    \includegraphics[width=1\columnwidth]{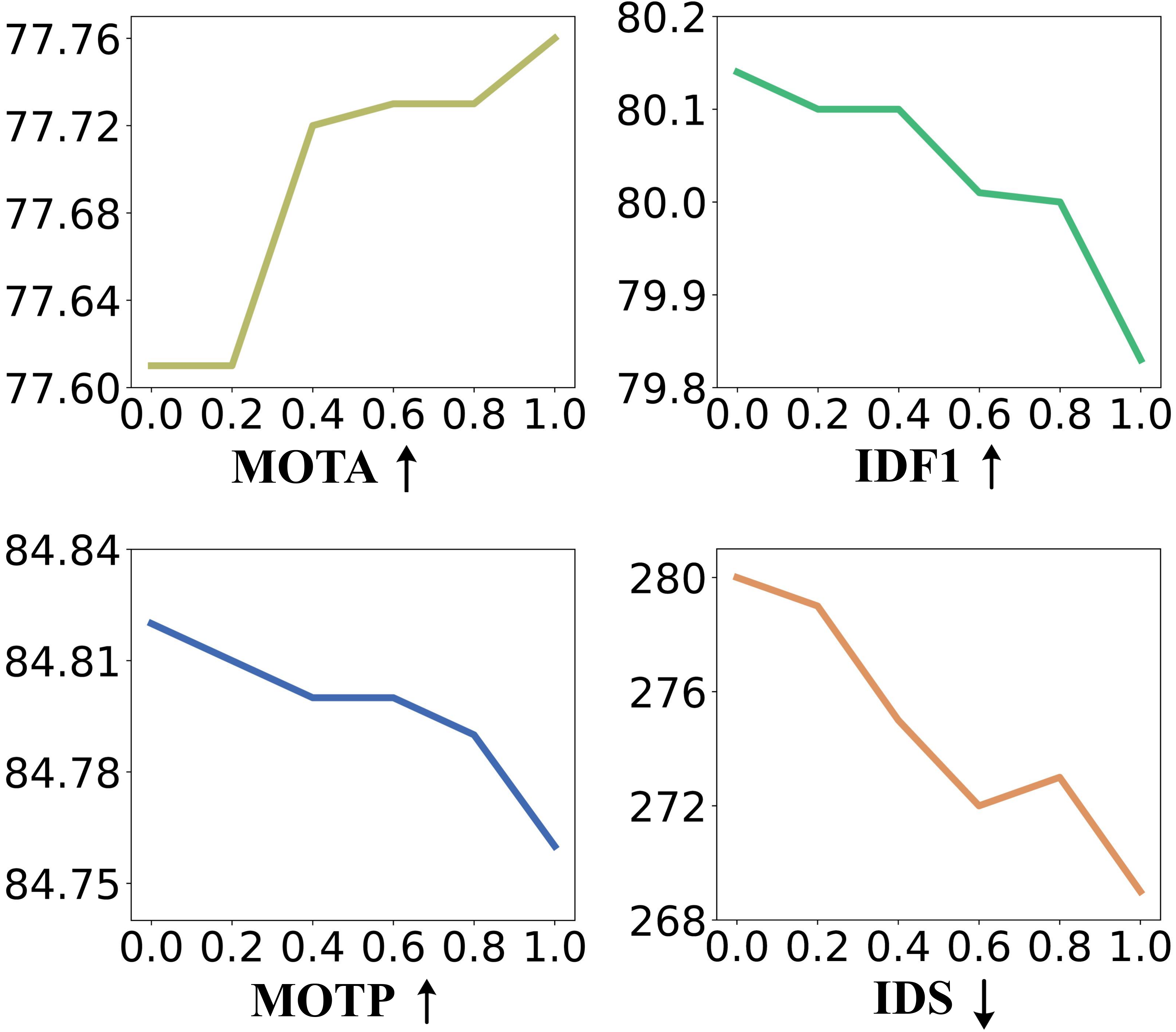}
    \caption{The impact of hyperparameter $\lambda$ on STRAM's performance.}
    \label{fig:STRAM_coefficient}
\end{figure}

The objective of STRAM is to seamlessly integrate both spatial and temporal regularization within associations. There exist diverse methods for achieving this integration. In our study, we adopted a straightforward yet useful technique involving a weighted summation of SRAM and TRAM. However, it's conceivable that employing more intricate fusion methods could yield even more promising results. Our experimentation extended to the exploration of various combination coefficient $\lambda$, as depicted in the Figure \ref{fig:STRAM_coefficient}. In order to strike a balance among all performance indicators, a coefficient of 0.5 was ultimately determined as the optimal choice.

\subsubsection{Association Stages}
ByteTrack \cite{zhang2022bytetrack} employs a two-stage association process, comprising a primary stage dedicated to linking high-confidence tracks with detections and a subsequent stage focusing on pairing residual tracks with low-confidence detections. As outlined by \cite{zhang2022bytetrack}, our configuration for ByteTrack establishes the detection score thresholds as $\tau_{high}=0.6$ and $\tau_{low}=0.1$.

Results obtained from the MOT17 dataset, employing two features across the two association stages, are presented in Table \ref{tab:ablation-study}. During the initial association stage, the inclusion of IoU+STRAM yields a noteworthy 0.29 increase in MOTA, a substantial 1.46 enhancement in IDF1, and an impressive reduction of 66 IDS instances when contrasted with utilizing IoU alone. Correspondingly, when Re-ID+STRAM is applied, there is a substantial boost of 1.98 in MOTA, an impressive 7.52 surge in IDF1, and a notable decrease of 77 IDS instances compared to utilizing Re-ID exclusively. These findings unequivocally affirm the potency of STRAM in enhancing association performance, given that the bulk of detections are associated in the primary stage. 

Moving on to the secondary association stage, despite STRAM receiving uncertain detections due to occlusion and motion blur, it adeptly generates appropriately aligned and complementary features. This is attested by the competitive performance of IoU+STRAM when juxtaposed with IoU alone and the superior performance of Re-ID+STRAM in comparison to Re-ID exclusive in this specific stage.

\begin{table} [tb!]
\renewcommand{\arraystretch}{1.2}
\centering 
\resizebox{1\columnwidth}{!}{
    \begin{tabular}{@{} *{2}l|*{3}l @{}}
    \toprule
    \textbf{Stage 1} & \textbf{Stage 2} & \textbf{IDF1} $\uparrow$ & \textbf{MOTA} $\uparrow$ & \textbf{IDS} $\downarrow$ \\
    
    \cmidrule{1-5}
    IoU & IoU & 79.55 & 77.65 & 333  \\
    IoU + STRAM  & IoU & \textbf{81.01(+1.46)} & 77.94(+0.29) & 267(-66)  \\
    IoU & IoU + STRAM & 79.70(+0.15) & 77.77(+0.12) & 328(-5)  \\
    IoU + STRAM & IoU + STRAM & 80.18(+0.63) & \textbf{78.00(+0.35)} & \textbf{245(-88)}  \\
    
    \cmidrule{1-5}
    Re-ID & Re-ID & 70.43 & 73.27 & 447  \\
    Re-ID + STRAM  & Re-ID & 77.95(+7.52) & 75.25(+1.98) & 370(-77)  \\
    Re-ID & Re-ID + STRAM & 71.73(+1.3) & 74.01(+0.74) & 409(-38)  \\
    Re-ID + STRAM & Re-ID + STRAM & \textbf{79.91(+9.48)} & \textbf{77.12(+3.85)} & \textbf{301(-146)}  \\
    \bottomrule
    \end{tabular}
}% end of scalebox
% \vspace{2mm}
\caption{Results of ByteTrack+STRAM across the two association stages on the MOT17 validation set}
\label{tab:ablation-study}
\end{table}

\subsection{More Comparison}

\subsubsection{Qualitative Comparison}

We conducted two qualitative experiments on the MOT17 dataset. One compared the results of ByteTrack \cite{zhang2022bytetrack} with and without RAM, while the other compared our RATracker with the typical rule-based method ByteTrack \cite{zhang2022bytetrack} and deep-learning based method DeepSORT \cite{wojke2017simple}. 

\begin{figure*}[tb!]
\centering\includegraphics[width=1.2\columnwidth]{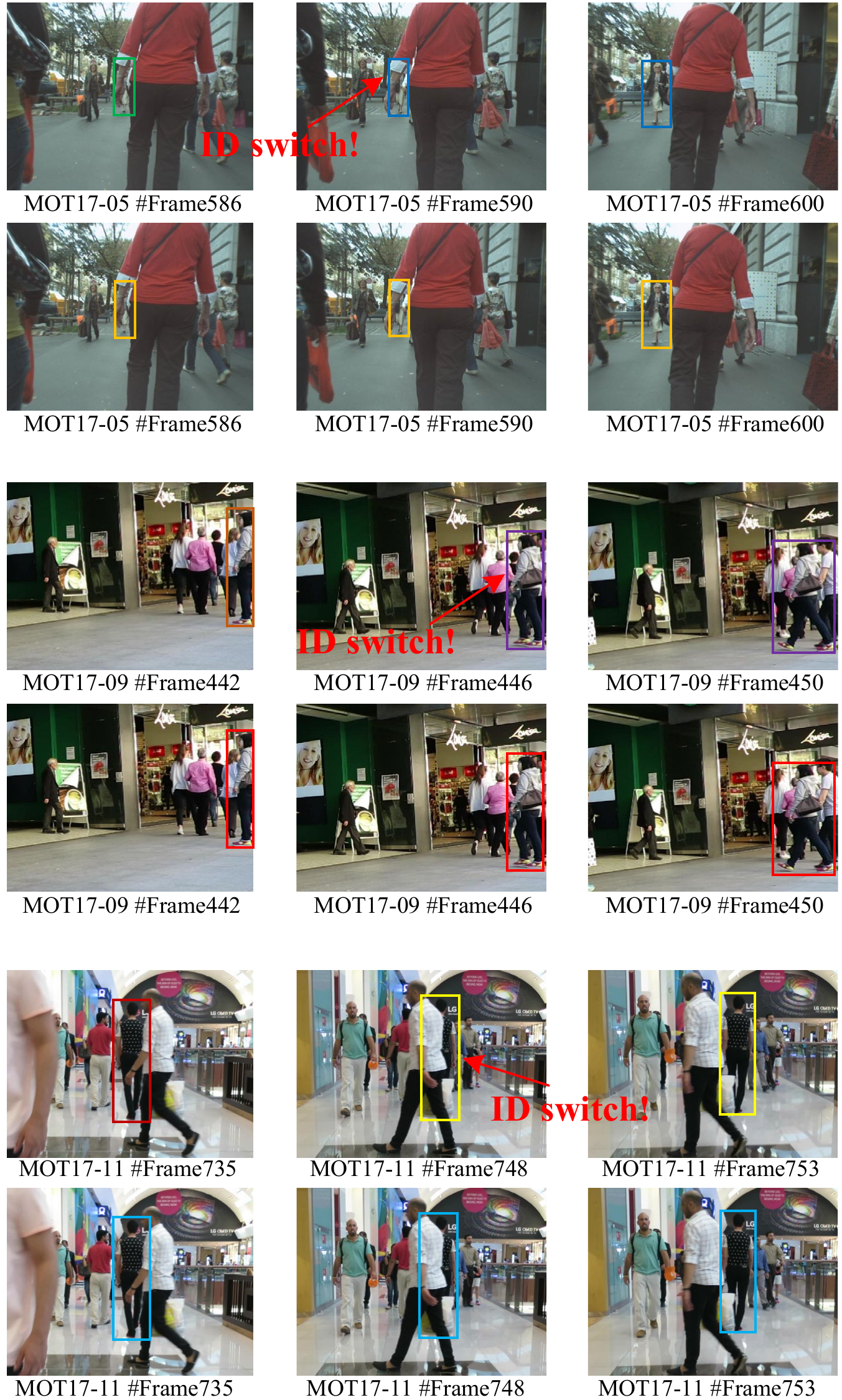}
\caption{ByteTrack vs. ByteTrack+STRAM. The upper rows feature ByteTrack's standalone tracking results, while the lower rows display the enhanced results from ByteTrack+STRAM. The ambient bounding boxes are deactivated for prioritizing target clarity.}\label{fig:vis}
\end{figure*}

The results of the first experiment conducted on three MOT17 scenarios, namely MOT17-05, MOT17-09, and MOT17-11, can be observed in Figure \ref{fig:vis}. The visualization showcases two sets of results: the upper rows depict the tracking results using ByteTrack alone, while the lower rows exhibit the results obtained through ByteTrack+STRAM. It's worth noting that ambient bounding boxes are disabled to enhance the visibility of the targets.

Across all these scenarios, instances of occlusion are prevalent. In the tracking results generated by ByteTrack on its own, there are instances where the issue of identity switching arises. However, this problem is effectively addressed in the tracking results produced by ByteTrack+STRAM. This indicates that the integration of RAMs holds promising potential for ensuring stable tracking performance, particularly in scenarios with occlusions.

\begin{figure}[tb]
    \centering
    \includegraphics[width=1\columnwidth]{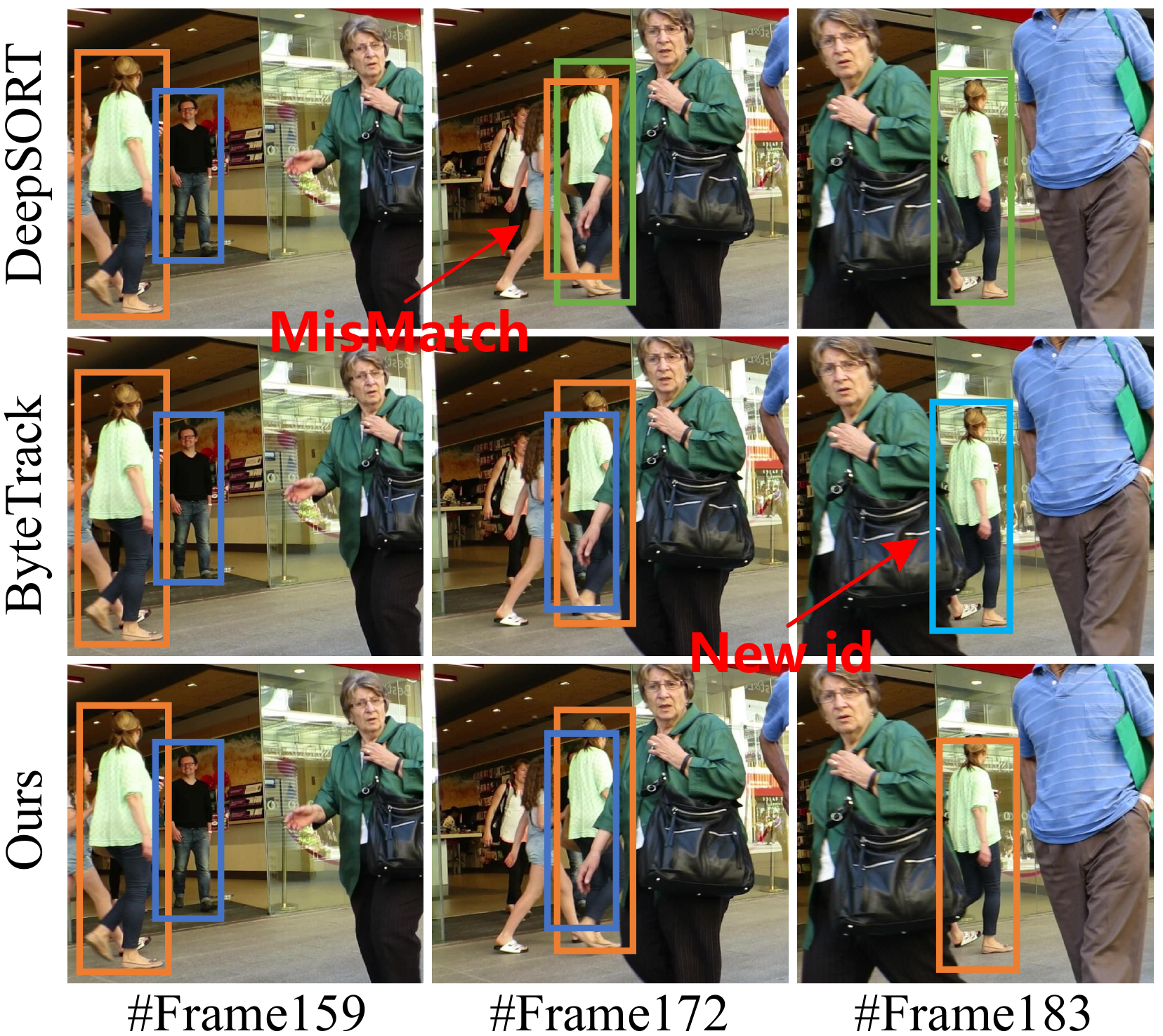}
    \caption{Comparison of rule-based, deep learning, and our methods on MOT17 validation set. Notable tracking errors emphasized.}
    \label{fig:qualitative_result}
\end{figure}

The findings from the second experiment are displayed in Figure \ref{fig:qualitative_result}. In scenarios where occlusion is notably prominent, both rule-based and deep learning-based methods face challenges in effectively establishing connections between targets before and after the occlusion. To ensure precise results, it becomes imperative to include additional temporal and spatial regularization on features, as showcased in our proposed approach.

\subsubsection{Feature Comparison}
We employ t-SNE as a visualization tool for Re-ID features extracted from targets within trajectories generated by RATracker. Our experiment focuses on two randomly selected scenes from the MOT17 dataset, where we choose 20 trajectories at random for visualization. The backbone tracker we utilize is JDE \cite{wang2020towards}. To assess the quality of clustering, we employ the Davies–Bouldin index (DBI) \cite{davies1979cluster}, where a lower DBI value indicates more cohesive clusters.

\begin{figure}[tb]
    \centering
    \includegraphics[width=1\columnwidth]{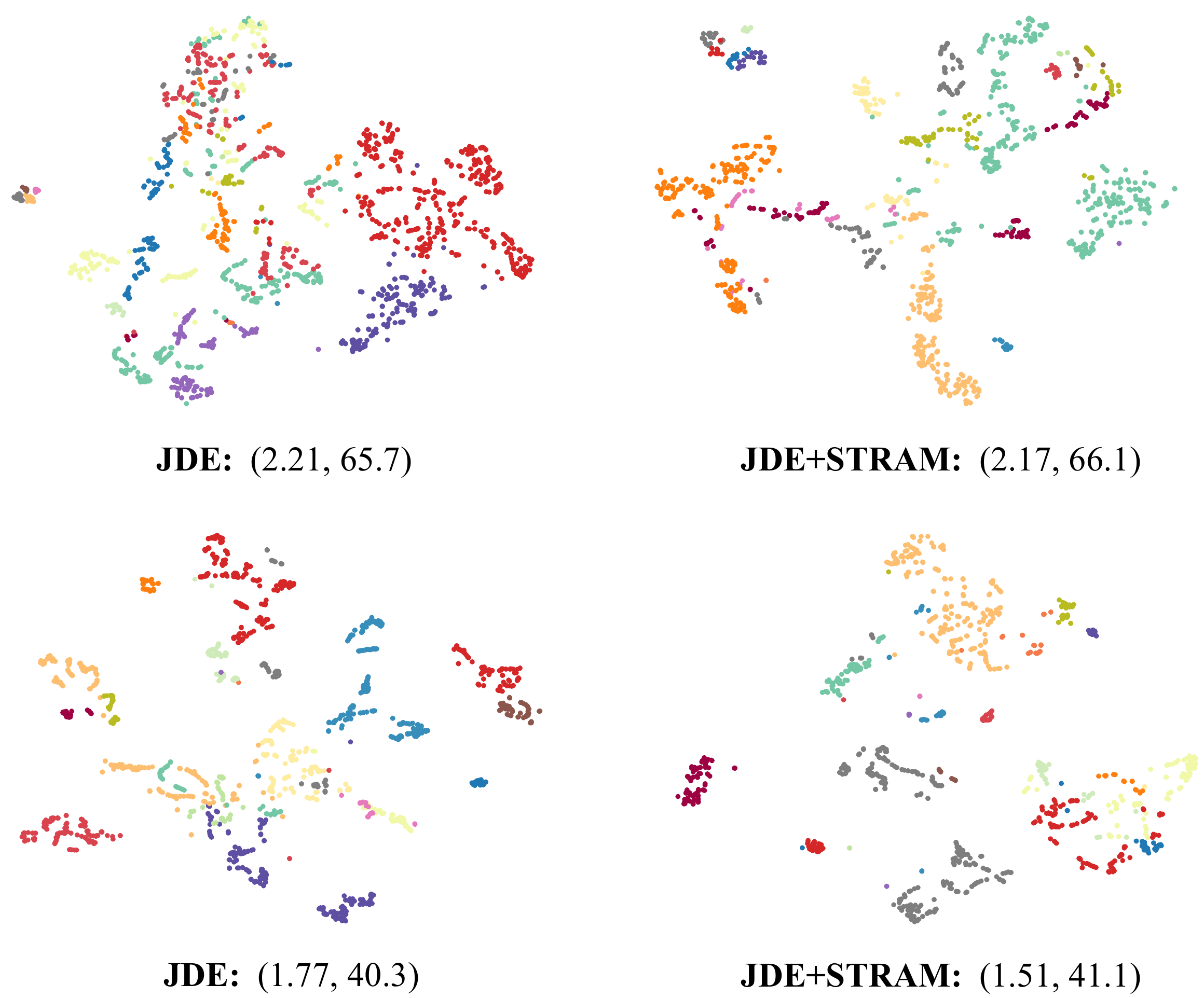}
    \caption{Visualizing Re-ID features via t-SNE: results on MOT17 scenes. Rows represent random scenes, while colors depict trajectories. Associated targets share the same color. Metrics (DBI$\downarrow$, MOTA$\uparrow$) shown in brackets alongside each row.}
    \label{fig:tsne}
\end{figure}

The visualization outcomes are depicted in Figure \ref{fig:tsne}. Each row in the figure represents results from a distinct random scene within MOT17. The colors represent individual trajectories, and points of the same color correspond to associated targets. Our observations reveal a notable distinction: points situated in the right column, generated by employing JDE+STRAM, exhibit greater clustering compared to those in the left column. This outcome is in alignment with the reduced DBI value attained by utilizing JDE+STRAM. Furthermore, the MOTA from association outcomes using JDE+STRAM surpasses that of using JDE alone. This improvement indicates that trajectories with more closely clustered targets bear a stronger resemblance to ground truth data. Evidently, this insight validates our adherence to representation alignment rules, highlighting the likelihood of targets within ground truth trajectories sharing similarities, thus reinforcing the robustness of our approach.

\subsubsection{Supervised vs Unsupervised}
Our RAM can undergo training using not only annotated data but also utilizing the real-time tracker's output to further enhance the tracker's performance during operation. Essentially, our method involves introducing constraints that ensure both temporal and spatial consistency during the tracking association phase. This is achieved through the contrastive regularization that benefits from the noise-resistant characteristics of the encoder training process, as discussed in the introduction section. This regularization incorporates a certain level of uncertainty, specifically derived from the output of the running tracker.

We validated this concept through an experiment. Initially, we executed the pre-trained ByteTrack once on the MOT17 validation dataset to obtain the initial tracking results. Subsequently, we trained the STRAM using triplets generated from these tracking results. We then evaluated the performance of ByteTrack combined with STRAM (ByteTrack+STRAM) on the same MOT17 validation dataset. For comparative purposes, we also trained another STRAM using triplets based on annotated bounding boxes.

\begin{figure}[tb]
    \centering
    \includegraphics[width=1\columnwidth]{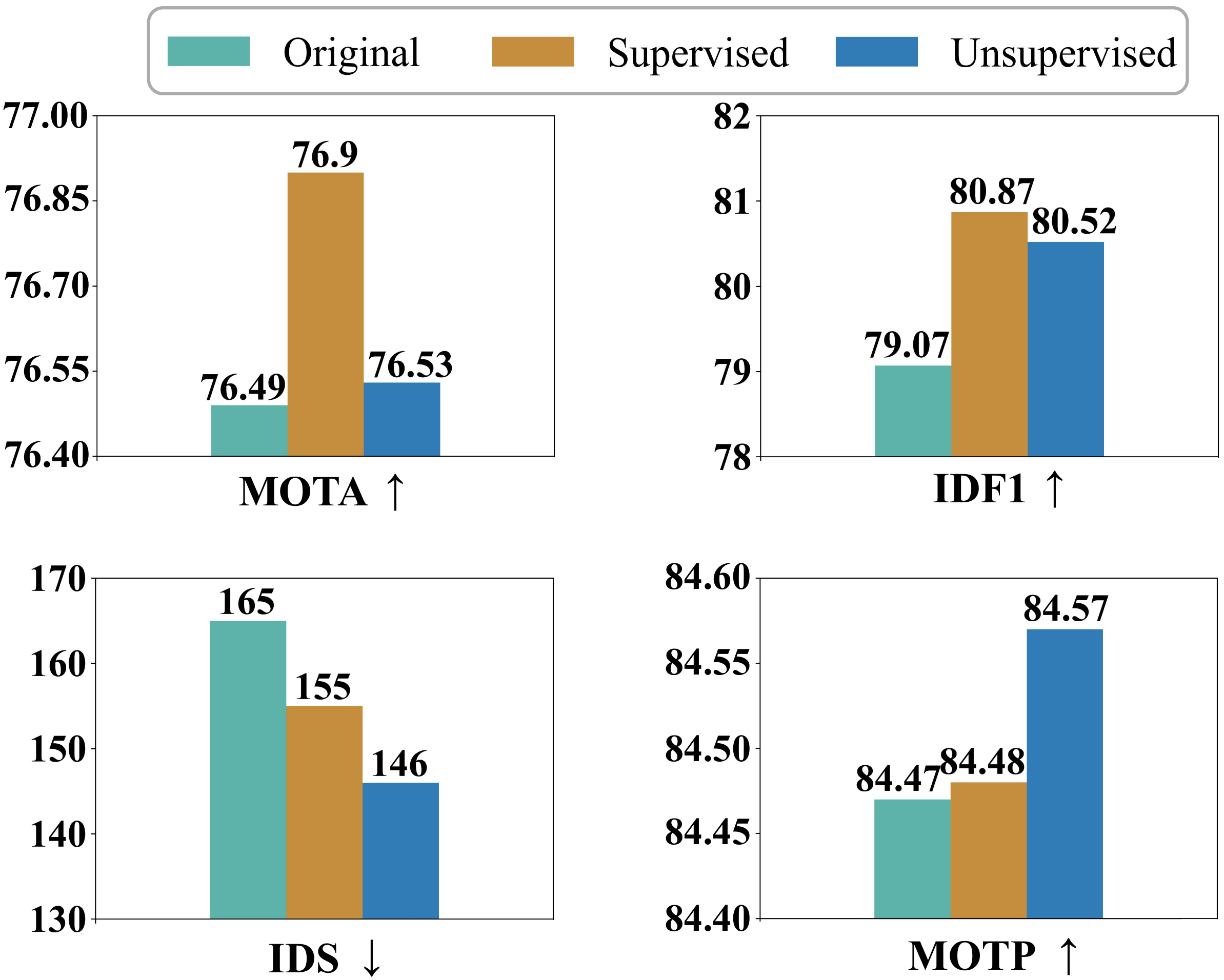}
    \caption{Results of STRAM trained by annotated boxes (supervised) and by the outputs of pretrained ByteTrack \cite{zhang2022bytetrack} (unsupervised) on the MOT17 validation set. }
    \label{fig:SupervisedVsUnsupervised}
\end{figure}

Figure \ref{fig:SupervisedVsUnsupervised} illustrates the performance comparison between STRAMs trained with annotated boxes (supervised) and those trained using the post refinement configuration (unsupervised). Although the unsupervised STRAM exhibits a slight decrease in performance compared to its supervised counterpart, it still surpasses the original tracker across all significant evaluation metrics.

\section{Conclusion} \label{sec:conc}
In this work, we have investigated two simple yet effective rules aimed at enhancing the MOT performance. These two rules encapsulate the concepts of spatial and temporal consistency among targets, acting as a form of contrasting regularization. Leveraging these rules, we have developed a streamlined encoding module termed RAM. This module serves to produce supplementary association features that complement the existing ones. Experiments conducted on the MOT17, MOT20 and BDD100K datasets have demonstrated that our proposed RAM is able to enhance the performance of various state-of-the-art trackers. Remarkably, this improvement persists even in scenarios where annotated data is not readily accessible.

\bmhead{Acknowledgments}
This research was supported by the National Natural Science Foundation of China under grant number 62002323.
% This research was funded by the National Natural Science Foundation of China grant number 62002323.

% \section*{Statements and Declarations}

% \begin{itemize}
% \item Competing Interests: The authors declare no conflict of interest. The funders had no role in the design of the study; in the collection, analyses, or interpretation of data; in the writing of the manuscript; or in the decision to publish the results.
% \end{itemize}

%\section*{Declarations}
%\begin{itemize} 
%
%\item Conflict of interest: The authors declare no conflict of interest. The funders had no role in the design of the study; in the collection, analyses, or interpretation of data; in the writing of the manuscript; or in the decision to publish the results.
%
%\item Availability of data: The dataset on MOT17 is available at \url{https://doi.org/10.48550/arXiv.1603.00831}.
%The dataset on MOT20 is available at \url{https://doi.org/10.48550/arXiv.2003.09003}.
%The dataset on BDD100K is available at \url{https://doi.org/10.48550/arXiv.1805.04687 }.
%
%% • Code availability: Code availability not applicable.
%
%% • Authors’ contributions:
%% XXXXXXXXX
%
%\end{itemize}
\bibliography{sn-bibliography}% common bib file

\end{document}